\documentclass[final]{article}

\usepackage{nips_2017arxiv}

\usepackage[utf8]{inputenc} 
\usepackage[T1]{fontenc}    
\usepackage{hyperref}       
\usepackage{url}            
\usepackage{booktabs}       
\usepackage{amsfonts}       
\usepackage{nicefrac}       
\usepackage{microtype}      

\usepackage{gensymb}
\usepackage{color}
\usepackage{amssymb,amsmath,mathtools}

\newcommand{\comment}[1]{}

\title{Improving Content-Invariance in Gated Autoencoders for 2D and 3D Object Rotation}

\author{Stefan Lattner\\
Austrian Research Institute for Artificial Intelligence\\
Vienna, Austria\\
\And
 Maarten Grachten\\
 Department of Computational Perception\\
 Johannes Kepler University\\
 Linz, Austria
}

\begin{document}

\maketitle

\begin{abstract}
Content-invariance in mapping codes learned by GAEs is a useful feature for various relation learning tasks.
In this paper we show that the content-invariance of mapping codes for images of 2D and 3D rotated objects can be substantially improved by extending the standard GAE loss (symmetric reconstruction error) with a regularization term that penalizes the symmetric cross-reconstruction error.
This error term involves reconstruction of pairs with mapping codes obtained from other pairs exhibiting similar transformations.
Although this would principally require knowledge of the transformations exhibited by training pairs, our experiments show that a bootstrapping approach can sidestep this issue, and that the regularization term can effectively be used in an unsupervised setting.
\end{abstract}

\section{Introduction}\label{sec:introduction}

Gated Autoencoders (GAEs) are a class of unsupervised neural network models especially suited for learning relations between data instances.
This capability is useful for tasks like facial expression modeling~\citep{susskind2011modeling}, motion learning~\citep{michalski2014modeling}, and perspective-invariant object recognition~\citep{ICML2012Memisevic_105}.
Another non-trivial task to which GAEs lend themselves is analogy-making, in which the objective is to produce a data instance X given the three instances A, B, C, and the query ``X is to A as B is to C''.

The aptness of GAEs to model such relations stems from three-way multiplicative (or ``gating'') connections between two input sources and a hidden mapping layer.
The advantage of this type of connections over additive connections (as used commonly in traditional neural networks) is that it allows for more independence between learned representations of content and transformation, respectively \citep{memisevic2013learning}.

In other words, an important factor contributing to the successful use of GAEs in relation learning tasks is that they learn representations of relations (we will refer to these representations as \emph{mapping codes}, see Section~\ref{sec:autoencoder}) that are to some degree invariant to the content in the inputs.
Given this observation, it is not far-fetched to ask whether models that are better at learning content-invariant mapping codes are also better at performing relation learning tasks.

In this paper we propose a novel training objective for GAEs, by extending the classical training objective with a \emph{content-invariance regularization} (CIR) term that penalizes what we call the \emph{cross-reconstruction error} (see Section~\ref{sec:regular}) between tuples of input pairs with similar mapping codes.
This approach leverages the fact that the standard GAE loss tends to encode relations between the inputs in the mapping codes, even if they are only weakly content-invariant.
Note that this bootstrapping procedure is completely unsupervised, and does not require labeled input pairs, where the type of transformation is known in advance.

We show experimentally that this approach leads to stronger content-invariance in representations of both 2D and 3D rotations of images, compared to standard GAEs.
First we compare the cross-reconstruction error of tuples of input image pairs conveying the same rotation angles.
We then assess the content-invariance of the mapping space in terms of cluster separation (see Section~\ref{sec:davies-bouldin-index}).
Thirdly, we use the mapping codes as inputs in a k-nearest neighbor classification task.
Furthermore, the effect of CIR is shown qualitatively using analogy making examples of rotated MNIST and NORB data.
Lastly and importantly, we find that CIR does not hinder the standard GAE loss.
To the contrary, it is often improved by the proposed regularization term.

The paper is organized as follows. Section \ref{sec:related-work} gives an overview on the usage of the GAE and discusses related models. In Section \ref{sec:method} the GAE is described and the CIR is introduced. The experiment setup including data preparation, training details and evaluation is described in Section \ref{sec:Experiment}. Results are presented and discussed in Section \ref{sec:results} and Section \ref{sec:concl-future-work} gives a conclusion.

\section{Related work}\label{sec:related-work}
GAEs utilize \emph{multiplicative interactions} to learn correlations between or within data instances. This principle was first used in cross-correlation models for motion in vision by \cite{adelson1985spatiotemporal} and later for binocular disparity by \cite{sanger1988stereo, qian1994computing, fleet1996neural}, based on hand-crafted Gabor filters \citep{memisevic2013learning}.

Using multiplicative interactions in combination with filter learning, \emph{higher-order} neural networks--computing sums over products--were proposed by \cite{rumelhart1986general, giles1987learning, smolensky1990tensor}, all of which were not directly related to vision.

\cite{tenenbaum2000separating} introduced \emph{bi-linear} models for separating person and pose of face images.
Bi-linear models are two-factor models whose outputs are linear in either factor when the other is held constant, a property which also applies to the GAE.
\cite{olshausen2007bilinear} proposed another variant of a bi-linear model in order to learn objects and their optical flow.
Due to its similar architecture, the Gated Boltzmann Machine (GBM) \citep{memisevic2007unsupervised, memisevic2010learning} can be seen as a direct predecessor of the GAE.
GBMs were applied to image pairs for learning transformations, for modeling facial expression \citep{susskind2011modeling}, and for disentangling facial pose and expression \citep{reed2014learning}. The GAE was introduced by \cite{memisevic2011gradient} as a derivative of the GBM, as standard learning criteria became applicable through the development of Denoising Autoencoders \citep{vincent2010stacked}.

While the before-mentioned contributions were mostly used to \emph{relate} different data instances, a series of publications exist which aim to capture within-image correlations based on the GBM \citep{hinton2010modeling, courville2011spike}, on the GAE \citep{memisevic2011gradient} and on other architectures \citep{wainwright1999scale, hoyer2002multi, hyvarinen2000emergence, karklin2006early}. These models are closely related to energy models \citep{hinton1981parallel}, based on the idea to square filter responses in a network.

GAEs were further used for analogy-making on rotated 3D objects and to learn transformation-invariant representations for classification tasks \citep{memisevic2013aperture, ICML2012Memisevic_105}, for parent-offspring resemblance \citep{dehghan2014look}, and for prediction of accelerated motion by stacking more layers in order to learn higher-order derivatives \citep{michalski2014modeling}.

In other domains, the GAE and some variants have been applied in linguistics to learn to negate adjectives \citep{rimell2017learning}, to activity recognition with the Kinekt sensor \citep{mocanu2015factored}, to robotics to learn to write numbers \citep{droniou2014learning} and to learning multi-modal mappings between action, sound, and visual stimuli \citep{droniou2015deep}.

\cite{cheung2014discovering} extended the common Autoencoder with a penalty term to disentangle object classes and their variations.

%
%

\section{Method}\label{sec:method}
In Section \ref{sec:autoencoder}, the GAE--as used in our experiments--is described, and Section \ref{sec:regular} introduces the Content-Invariance Regularization (CIR).

\begin{figure}
\begin{center}
\includegraphics[width=.4\linewidth]{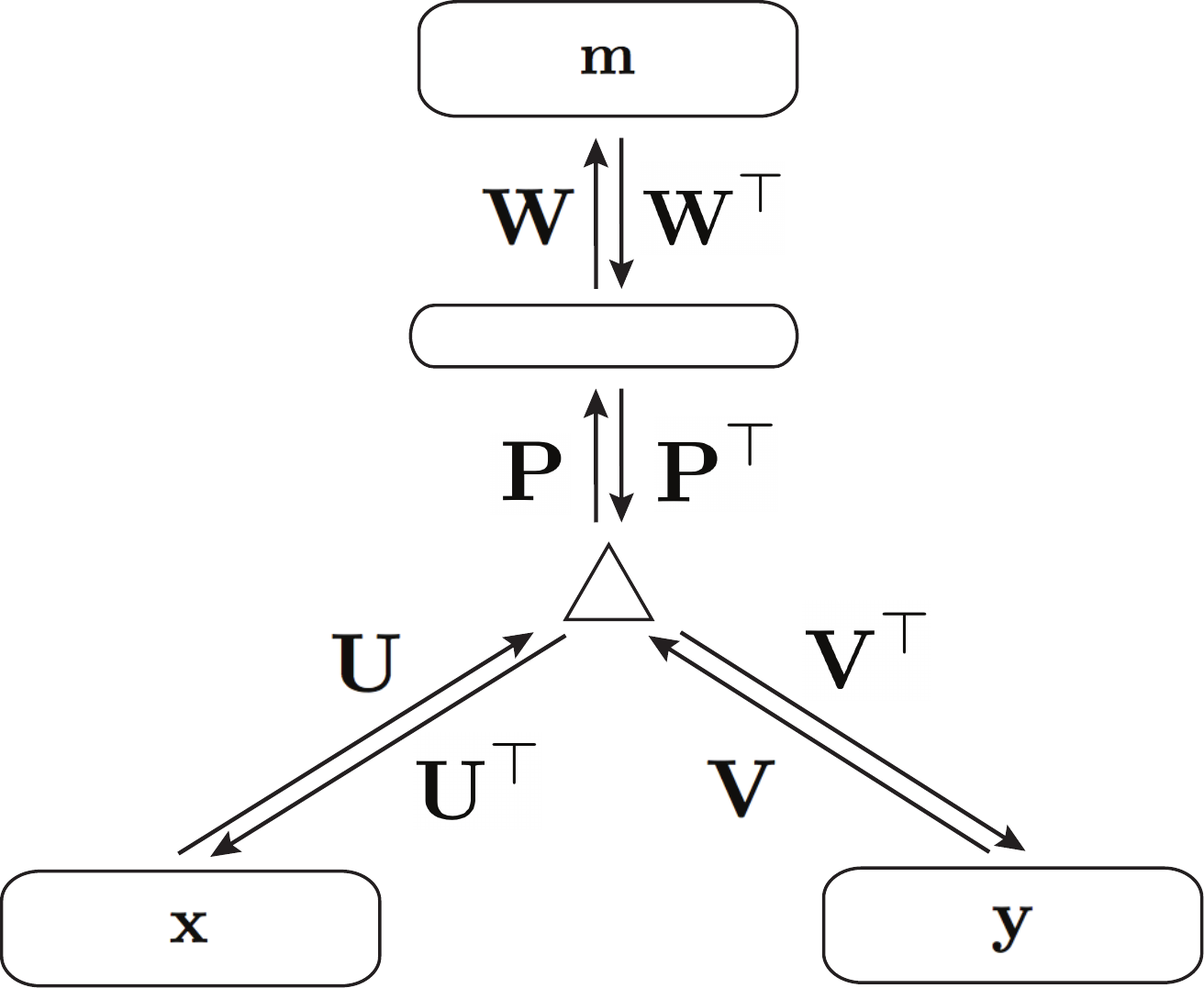}
\caption{Schematic illustration of a Gated Autoencoder.}
\label{fig:ga}
\end{center}
\end{figure}

\subsection{Gated Autoencoder} \label{sec:autoencoder}
Given a set of data pairs $(\mathbf{X},\mathbf{Y})$
with pairs $\mathbf{x}_i,\mathbf{y}_i\in\mathbb{R}^{P}, i = 0 \dots N$,
where each part of pair $i$ is a \emph{transformed} version of the other.
A Gated Autoencoder (GAE) can learn such transformations, representing them in so-called mapping units as \emph{mapping codes} $\mathbf{m}_i\in\mathbb{R}^{L}$, computed as
\begin{equation}\label{eq:gamap}
\mathbf{m}_i = \sigma (\mathbf{W}\mathbf{P}(\mathbf{U}\mathbf{x}_i \cdot \mathbf{Vy}_i)),
\end{equation}
where $\mathbf{U}, \mathbf{V} \in \mathbb{R}^{P \times O}$ and $\mathbf{W} \in \mathbb{R}^{O \times L}$ are weight matrices, $\sigma$ is a non-linearity, the operator $\cdot$ denotes the element-wise multiplication, and $\mathbf{P}$ is a band-diagonal matrix (see Figure \ref{fig:ga} for a depiction of the GAE):
\begin{equation}
\mathbf{P}=\begin{pmatrix}
1 & 1 & 0 & 0 &\cdots & 0 & 0\\
0 & 0 & 1 & 1 & \cdots & 0 & 0\\
\vdots & \vdots &  \vdots & \vdots &    &    & \vdots\\
0 & 0 & 0 & 0 & \cdots & 1 & 1\\
\end{pmatrix}
\end{equation}

In a trained GAE, $\mathbf{U}$ and $\mathbf{V}$ contain a set of filter pairs constituting the eigenvectors of the training data before and after a transformation.
Projecting inputs $\mathbf{x}$ and $\mathbf{y}$ onto the filter pairs results in a decomposition of the data into subspaces.
When $\mathbf{U}$ contains the real parts and $\mathbf{V}$ contains the imaginary parts of the eigenvectors,
a specific transformation between $\mathbf{x}$ and $\mathbf{y}$ amounts to a number of concurrent subspace rotations in the complex plane.
Therefore, the element-wise (i.e.
pairwise) multiplication of filter responses in Equation \ref{eq:gamap} results in \emph{rotation detectors} which we refer to as ``factors''.

Pooling two neighboring factors with $\mathbf{P}$ is an extension to the common GAE (as proposed by \cite{memisevic2013aperture}), explicitly separating within-subspace pooling from across-subspace pooling (in the common GAE performed by $\mathbf{W}$ only), reducing the complexity in learning $\mathbf{W}$.
Using $\mathbf{P}$ forces all four corresponding filters to remain in the same subspace, where the two filters within each weight matrix are approximately in quadrature.
As a result, the factors after pooling representing relative rotation angles are less sensitive to absolute starting angles.
Matrix $\mathbf{W}$ pools \emph{across} subspaces, causing mapping units to learn common transformations in the data.
For more details on that, please refer to \cite{ICML2012Memisevic_105}.

An instance can be reconstructed given the other instance and a mapping code as
\begin{equation}\label{reconx}
\mathbf{\tilde{x}}_i = \mathbf{U^\top} (\mathbf{P}^\top\mathbf{W}^\top \mathbf{m}_i \cdot \mathbf{Vy}_i),
\end{equation}
and likewise
\begin{equation}\label{recony}
\mathbf{\tilde{y}}_i = \mathbf{V}^\top (\mathbf{P}^\top\mathbf{W}^\top \mathbf{m}_i \cdot \mathbf{Ux}_i).
\end{equation}
A common way to train a GAE \citep{memisevic2013learning,michalski2014modeling} is to minimize the symmetric reconstruction error
\begin{equation}\label{eq:recon_symm}
\mathcal{L}_{\text{sre}} = \|\mathbf{x}_i - \mathbf{\tilde{x}}_i\|^2 + \|\mathbf{y}_i - \mathbf{\tilde{y}}_i \|^2.
\end{equation}

\subsection{Content-Invariance Regularization}\label{sec:regular}
\newcommand{\fatm}{\ensuremath{\mathbf{m}}}
\newcommand{\fatx}{\ensuremath{\mathbf{x}}}
\newcommand{\faty}{\ensuremath{\mathbf{y}}}

The symmetric reconstruction error cost function (cf. Equation \ref{eq:recon_symm}) penalizes the reconstruction error in both the forward and backward transformations for each training pair.
However, this does not guarantee a mapping $\mathbf{m}_i$, encoding a specific transformation, results in minimal reconstruction error for \emph{all pairs} exhibiting that transformation.
If we would know the transformation exhibited by pairs in the training data, this objective could be easily enforced by selecting pairs with the same transformation (in case of discrete transformation types), or similar transformations (in case of continuous transformation types), and minimizing the reconstruction error of one pair given the mapping code of the other pair. We call this the \emph{cross-reconstruction error}. 
In the absence of this knowledge, we don't know a priori which pairs should be selected for minimizing the cross-reconstruction error.

However, if we assume that variation in transformation accounts for more variance in the mapping space than variation in content does, it follows that codes that are close in the mapping space are more likely to encode similar transformations than similar content.
In that case, reducing the \emph{cross-reconstruction error} between pairs that are close to each other in the mapping space will increase the content-invariance of the mapping codes.
This assumption can of course not be expected to hold in the initial stages of training a GAE, but becomes increasingly likely as the symmetric reconstruction error (the standard GAE loss) decreases during training.




This leads us to the following formulation of content-invariance regularization (CIR).
Let $\mathbf{M}\in\mathbb{R}^{N\times L}$ denote the mapping codes computed for training pairs $(\mathbf{X},\mathbf{Y})$, such that $\mathbf{x}_i,\mathbf{y}_i$ is the data pair yielding code $\mathbf{m}_i$.
For a given pair $i$, let $\mathbf{m}_{(1)} \dots \mathbf{m}_{(N-1)}$ be a reordering of $\mathbf{M}/\fatm_i$ such that
\begin{equation}
\|\mathbf{m}_i-\mathbf{m}_{(1)}\| \leq \dots \leq \|\mathbf{m}_i-\mathbf{m}_{(N-1)}\|.
\end{equation}
Then, given a randomly chosen random code $\mathbf{m}_j$ from the $k$\footnote{$k$ is a hyperparameter of the regularization term, to be chosen in advance or adapted according to some scheme. See Section~\ref{sec:training-details}.} nearest neighbors $\mathbf{m}_{(1)} \dots \mathbf{m}_{(k)}$ of $\mathbf{m}_i$, $\mathbf{x}_i$ and $\mathbf{y}_i$ are cross-reconstructed as
\begin{equation}
\mathbf{\widetilde{x}}_i' = \sigma (\mathbf{U^\top} (\mathbf{P}^\top\mathbf{W}^\top \mathbf{m}_j \cdot \mathbf{Vy}_i)),
\end{equation}
and
\begin{equation}
\mathbf{\widetilde{y}}_i' = \sigma (\mathbf{V}^\top (\mathbf{P}^\top\mathbf{W}^\top \mathbf{m}_j \cdot \mathbf{Ux}_i)),
\end{equation}
respectively.
The \emph{symmetric cross-reconstruction error} is the error between the known instances and their cross-reconstructions:
\begin{equation}\label{eq:recon_symm_reg}
\mathcal{L}_{\text{scre}} = \|\mathbf{x}_i - \mathbf{\widetilde{x}}_i' \|^2 + \|\mathbf{y}_i - \mathbf{\widetilde{y}}_i'\|^2.
\end{equation}

Finally, the symmetric cross-reconstruction error is linearly interpolated with the original symmetric reconstruction error (cf. Equation \ref{eq:recon_symm}) as
\begin{equation}\label{eq:recon_symm_reg}
\mathcal{L} = (1- \lambda) \mathcal{L}_{\text{sre}} + \lambda \, \mathcal{L}_{\text{scre}},
\end{equation}
where the scalar $\lambda$ controls the strength of the regularization, and is intended as increasing from 0 to 1 over the course of training, according to some scheme.
We will return to this issue in Section~\ref{sec:training-details}.

%
%

\section{Experiment}\label{sec:Experiment}
The invariance-regularization is tested quantitatively and qualitatively, on different data sets and for different combinations of training and evaluation data.
Section \ref{sec:data} introduces the used datasets and data preparation, Section \ref{sec:training-details} gives training details, and the evaluation of the method is described in Section \ref{sec:training-details}.

\subsection{Data}\label{sec:data}
The majority of tests is performed on rotated MNIST digits \citep{lecun2010mnist}.
We use the train set, validation set, and test set split of the original MNIST data set, and thereof we create different transformation sets.
The MNISTR20 transformation set includes pairs of $\{-180,-160,-140,\dots,160\}$ degree rotation angle, resulting in $18$ discrete transformation classes.
For the MNISTR20/10 set, all angles of the MNISTR20 set are increased by $10$ degrees, in order to obtain the complementary classes $\{-170,-150,\dots,170\}$.
The MNISTR1 transformation set include pairs whose mutual rotations exhibit all possible discrete angles of $[-180,179]$ degrees, resulting in $360$ classes.

For the experiments on rotated 3D objects, the train set of the small NORB dataset\footnote{\url{http://www.cs.nyu.edu/~ylclab/data/norb-v1.0-small/}} is used, which consists of videos of rotating objects in 5 categories (e.g. four-legged animals, cars, planes), with 9 objects in each category. For training the first eight objects, and for testing the ninth objects in each category are used. From the videos, pairwise frames with an azimuth angle of [-20, 20] degrees, with fixed elevation of 30 degrees and fixed lighting condition are extracted.

All datasets are contrast-normalized to zero mean and unit variance, a crucial preparation for usage with the GAE \citep{memisevic2013aperture}.

\subsection{Training Details}\label{sec:training-details}
The CIR takes a bootstrapping approach relying on a pre-training of the model.
Therefore, the strength $\lambda$ and the number of nearest neighbors $k$ to swap mapping codes with (cf. Section \ref{sec:regular}) should be minimal at the beginning of the training.
A scheme which works well for the MNIST data is a linear increase of the parameters during training, where we increase both parameters to their maximum values $\lambda = 1$ and $k=10$ over $3000$ epochs.
For the NORB dataset, a stepwise increase of the CIR parameters has shown to work best.

The GAE is trained using stochastic gradient descent, where $50\%$ of the input units are turned off at random.
Enforcing sparsity on the mapping units and on the factors, as well as weight regularization is crucial for a good performance of the model.
A further improvement is reached by limiting the weight norms and by clipping the gradients during training.
A cost term is used which penalizes the deviation of the norm of each input filter to the average filter norm, as well as the deviation from zero mean.

\subsection{Evaluation}\label{sec:evaluation}
For all experiments, the GAE is trained on different \emph{GAE Data} sets, and tested using different \emph{Eval Data} sets.
The Mean Symmetric Reconstruction Error (MSRE) and the Mean Symmetric Cross-Reconstruction Error (MSCRE) are measured by reconstructing the test set of the respective Eval Data using the GAE trained on the training set of the respective GAE Data.
The Davies-Bouldin Index (DBI) (see Section \ref{sec:davies-bouldin-index}) and the Rotation Error (see Section \ref{sec:rot-error}) are evaluated based on the mapping codes inferred by projecting the test set of the respective Eval Data in the mapping space of the GAE trained on the training set of the respective GAE Data.

\subsubsection{Davies-Bouldin Index}\label{sec:davies-bouldin-index}
The Davies-Bouldin Index (\textit{DB}) \citep{davies1979cluster} is a metric for the quality of the clustering of data points. It is commonly used to evaluate clustering algorithms and their parameters. For example, in $k$-means clustering \citep{hartigan1975clustering}, the \textit{DB} can be used to determine $k$. \textit{DB} amounts to the average ratio between the scatter within clusters and the distance between cluster centers. It is minimal, when clusters are of minimal expansion and of maximal mutual distance. When content information does not contribute to the variance in the content-invariant representations of transformations, resulting clusters are of smaller expansion. Therefore, we use the \textit{DB} as a metric for evaluating the quality of the representations learned by the GAE.

\subsubsection{Rotation Error}\label{sec:rot-error}
The error of the GAEs predicted rotation angles for the Eval Data is evaluated as follows.
A K-Nearest Neighbor (KNN) classifier is used to predict the discrete rotation angles for the test set pairs of the Eval Data.
From these predictions, the mean distances to the \emph{closest} true rotation angles is calculated.
For example, when the predicted rotation angle for a test pair is $-175$ degrees, and the true angle is $170$ degrees, that pair contributes an error of $15$ degrees.

\section{Results and Discussion}\label{sec:results}
The quantitative results from the experiments are shown in Table \ref{results_recon}. The GAE with the content-invariance regularization (CIR) clearly outperforms the GAE without CIR on almost all combinations of GAE Data and Eval Data.
In particular, it is interesting to note that the MSRE is also reduced (in all but one case) when using the GAE with CIR.
This is an unexpected result, since the training of the GAE without CIR optimizes solely MSRE, rather than MSRE and MSCRE jointly.
This suggest that by reducing content-variance in the mapping space, CIR facilitates generalization over transformations.
The better MSCRE on the MNISTr20/10 Eval Data set supports this assumption, because none of these rotation angles were in the training set.
This comes at the cost of slightly higher MSRE on that set indicating that a higher generalization on transformations results in lower specificity on content types.
The lower DBI values of GAE+CIR with respect to GAE show that the clusters formed by pairs with identical transformations (e.g. all instances of numbers rotated by 60 degrees) are more compact in the GAE+CIR.

\begin{table}[t]
\vspace{-.27cm}
\centering
\footnotesize
\begin{tabular}{rlllllll}
\toprule
GAE Data & Eval Data & MSRE & MSCRE & DBI & Rot. Err (deg.)\\ \midrule
\multicolumn{1}{l}{\textbf{GAE \rule{0pt}{2ex}}\ \ \ \ \ \ \ \ } & & & & & &\\
MNISTr20 & MNISTr20 & 0.0146 & 0.0216 & 0.100 & 0.198\\
MNISTr20 & MNISTr20/10 & \textbf{0.0178} & 0.0250 & 0.122 & 0.344\\
MNISTr20 & MNISTr1 & 0.0162 & 0.0233 & 0.953 & 1.862\\
\rule{0pt}{2.5ex}
MNISTr1 & MNISTr20 & 0.0174 & 0.0194 & 0.086 & 0.180\\
MNISTr1 & MNISTr20/10 & 0.0173 & 0.0194 & 0.085 & 0.218\\
MNISTr1 & MNISTr1 & 0.0173 & 0.0194 & 0.970 & 1.206\\
\rule{0pt}{2.5ex}
NORB & NORB & 0.0067 & 0.0241 & 2.351 & 0.650 \\
\multicolumn{1}{l}{\textbf{GAE+CIR} \rule{0pt}{2.5ex}} & & & & & &\\
MNISTr20 & MNISTr20 & \textbf{0.0130} & \textbf{0.0149} & \textbf{0.071} & \textbf{0.054}\\
MNISTr20 & MNISTr20/10 & 0.0193 & \textbf{0.0213} & \textbf{0.077} & \textbf{0.128}\\
MNISTr20 & MNISTr1 & \textbf{0.0161} & \textbf{0.0180} & \textbf{0.873} & \textbf{0.920}\\
\rule{0pt}{2.5ex}
MNISTr1 & MNISTr20 & \textbf{0.0154} & \textbf{0.0171} & \textbf{0.082} & \textbf{0.162}\\
MNISTr1 & MNISTr20/10 & \textbf{0.0153} & \textbf{0.0170} & \textbf{0.084} & \textbf{0.146}\\
MNISTr1 & MNISTr1 & \textbf{0.0154} & \textbf{0.0171} & \textbf{0.930} & \textbf{0.954}\\
\rule{0pt}{2.5ex}
NORB & NORB & \textbf{0.0038} & \textbf{0.0061} & \textbf{0.174} & \textbf{0.323}\\
\bottomrule
\end{tabular}
\vspace{0.1cm}
\caption{Results for the Gated Autoencoder with (GAE+CIR) and without (GAE) content-invariance regularization, on different datasets. Section~\ref{sec:data} describes the referenced datasets, and Section~\ref{sec:evaluation} describes the evaluation criteria. The column \emph{GAE Data} specifies the dataset used to train the GAE, and the column \emph{Eval Data} shows the dataset used for evaluation.}
\label{results_recon}
\end{table}

\begin{figure}
\begin{center}
\includegraphics[width=1.\linewidth]{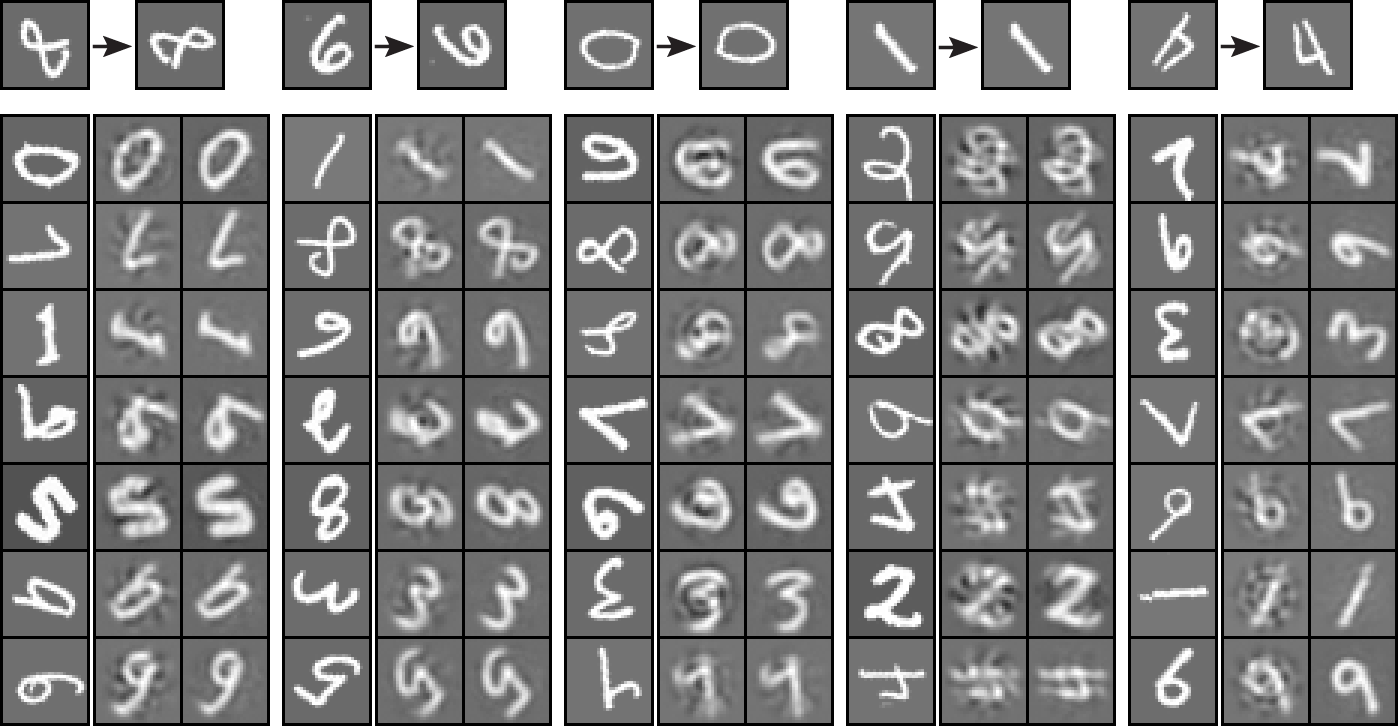}
\caption{Analogies for rotated MNIST digits. Mappings are inferred from the pairs in the upper-most row and applied on the images of the first column of each group. The second column of each group shows analogies generated by the vanilla GAE, the right-most columns show analogies rendered by the GAE using CIR.}
\label{fig:analogies_mnist}
\end{center}
\end{figure}

Figure \ref{fig:analogies_mnist} shows analogy making examples for the MNISTr20 dataset. The better quality of the analogies using the invariance-regularization indicates that the mappings learned by the GAE better represent the content-invariant transformations. The fourth group shows an ambiguous transformation, as the ``1'' in the input may represent both a rotation by $0$ and by $180$ degrees. This ambiguity is visible in the transformation results: they show two superimposed rotations (particularly noticeable by the two loops in the reconstruction of the ``2'' in the first row).
While the GAE+CIR can obviously not resolve this ambiguity, its reconstruction is less noisy.

\begin{figure}
\begin{center}
\includegraphics[width=1.\linewidth]{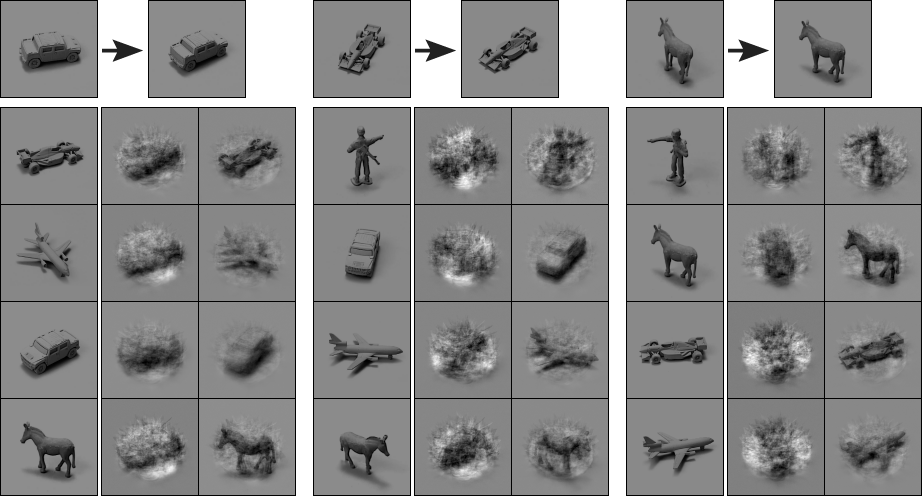}
\caption{Analogies for 2-frame NORB videos. Mappings are inferred from the pairs in the upper-most row and applied to the images of the first column of each group. The second column of each group shows analogies generated by the vanilla GAE, the right-most columns show analogies rendered by the GAE using CIR.}
\label{fig:analogies_norb}
\end{center}
\end{figure}

Figure \ref{fig:analogies_norb} shows analogy making examples for the NORB dataset similar to \cite{memisevic2013aperture}, who used a variant of the standard GAE with concatenated video frames, tied weights and ``gating noise''.
Even with this variant, it was not always possible to apply transformations inferred from image pairs in an analogy making task (some inferred rotations lead to noisy non-transformed copies of the source images and, remarkably, the identity transformation sometimes leads to a rotation in the analogy renderings).
As our experiments show, with the vanilla GAE \citep{memisevic2011gradient}, it is almost impossible to solve this task.
Only source images very similar to the query source can be approximately transformed.
The CIR mitigates this problem, making it possible to learn content-invariant transformations with the vanilla GAE.

\section{Conclusion and future work}\label{sec:concl-future-work}
The empirical results reported above show that while GAEs are able to learn mapping codes to transformations that are to some degree content-invariant (especially in 2D image rotations), the content-invariance of the mapping codes can be substantially improved by including a regularization term that penalizes the cross-reconstruction error.

Although this approach principally requires knowledge of the transformations exhibited by data pairs (cross-reconstruction error should only be penalized for pairs with similar transformations), we show that this knowledge is not necessary in practice, and that a bootstrapping approach in which the influence of the regularization term is incrementally increased during training allows it to work in unsupervised settings, by relying on the assumption that similar mapping codes represent similar transformations.

Interestingly, the results also show that in many cases training with content-invariance regularization also leads to lower standard GAE loss (symmetric reconstruction error), despite the fact that the content-invariance regularization competes with the reconstruction error during training.

In this paper we have limited our experiments to learning 2D and 3D object rotations in images.
The regularization method is obviously not limited to this type of data and transformations, and future work should assess the merits of the proposed method on other data, and using different transformations.

\section*{Acknowledgments}
This work is supported by the European Research Council (ERC) under the EU's Horizon 2020 Framework Programme (ERC Grant Agreement number 670035, project CON ESPRESSIONE).

\small
\bibliographystyle{named}
\bibliography{bib/bib_mg,bib/bib_cc,bib/bib_sl}

\end{document}